\documentclass[11pt]{article}

\usepackage[preprint]{acl}

\usepackage{times}
\usepackage{latexsym}

\usepackage[T1]{fontenc}
\usepackage[utf8]{inputenc}

\usepackage{microtype}

\usepackage{inconsolata}

\usepackage{graphicx}

\usepackage{booktabs}
\usepackage{multirow}

\title{Mitigating Prompt-Induced Hallucinations in Large Language Models via Structured Reasoning}

\author{Jinbo Hao\textsuperscript{1*}, Kai Yang\textsuperscript{2*}, Qingzhen Su\textsuperscript{1}, Yang Chen\textsuperscript{2}, Yifan Li\textsuperscript{2}, Chao Jiang\textsuperscript{2} \\
$^1$School of Computer Engineering, Jiangsu Ocean University\\ $^2$School of Computer Science and Technology, Soochow University\\
}

\begin{document}
\maketitle
\begin{abstract}
To address hallucination issues in large language models (LLMs), this paper proposes a method for mitigating prompt-induced hallucinations. Building on a knowledge distillation chain-style model, we introduce a code module to guide knowledge-graph exploration and incorporate code as part of the chain-of-thought prompt, forming an external knowledge input that provides more accurate and structured information to the model. Based on this design, we develop an improved knowledge distillation chain-style model and leverage it to analyze and constrain the reasoning process of LLMs, thereby improving inference accuracy. We empirically evaluate the proposed approach using GPT-4 and LLaMA~3.3 on multiple public datasets.
Experimental results demonstrate that incorporating code modules significantly enhances the model’s ability to capture contextual information and effectively mitigates prompt-induced hallucinations. Specifically, HIT@1, HIT@3, and HIT@5 improve by 15.64\%, 13.38\%, and 13.28\%, respectively. Moreover, the proposed method achieves HIT@1, HIT@3, and HIT@5 scores exceeding 95\% across several evaluation settings. These results indicate that the proposed approach substantially reduces hallucination behavior while improving the accuracy and verifiability of large language models.

\end{abstract}

\section{Introduction}

Large language models (LLMs) acquire contextual linguistic relationships by learning from massive-scale data, enabling them to model language semantics and to perform tasks such as language understanding, text generation, machine translation, and question answering \cite{brown2020language,raffel2020exploring}. These capabilities allow LLMs to support the translation and generation of multimodal content, including speech, text, images, and videos, and have led to their widespread adoption across numerous natural language processing applications \cite{bommasani2021opportunities}. Recent studies further demonstrate that LLMs can be extended beyond surface-level language processing to support structured reasoning tasks \cite{yang2024harnessing, yang2024can, xiong2024large}.

However, the fundamental mechanism of LLMs relies on probabilistic prediction learned from training data. Because the training data inevitably contains noise, biases, and incomplete knowledge, LLMs may generate responses that are fluent but factually incorrect or logically inconsistent \cite{maynez2020faithfulness}. This phenomenon is commonly referred to as hallucination. In particular, prompt-induced hallucinations arise when the model produces erroneous outputs due to ambiguous, incomplete, or misleading prompts, even when the underlying task is well-defined \cite{ji2023survey, tonmoy2024comprehensive}.

\begin{figure*}[t]
  \centering
  \includegraphics[width=0.8\linewidth]{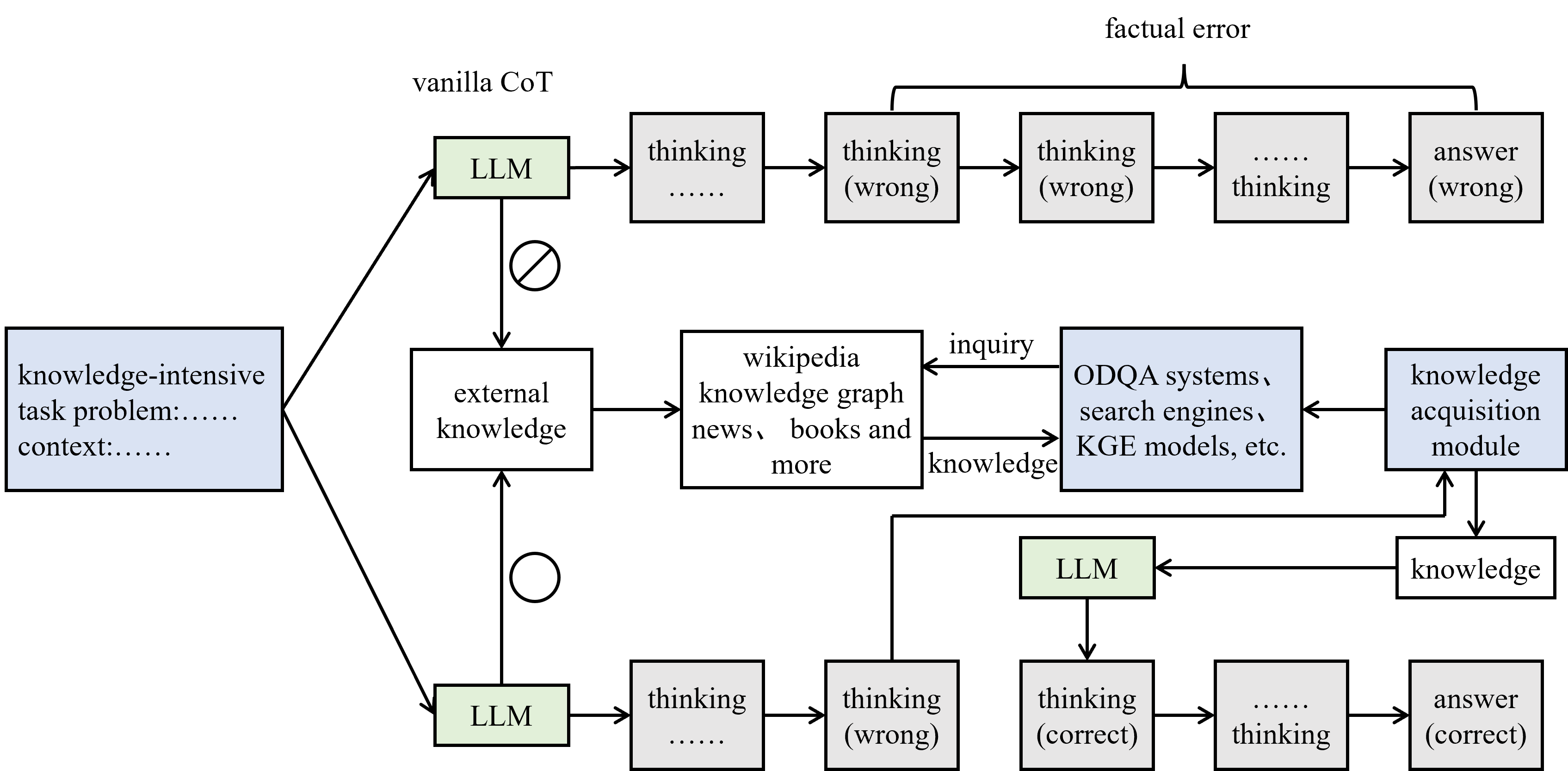}
  \caption{Structure of the knowledge distillation chain model.}
\end{figure*}

Hallucinations significantly limit the reliability and practical deployment of LLMs in high-stakes domains such as scientific research, medical decision-making, and legal analysis \cite{bender2021dangers}. Existing approaches to mitigating hallucinations include improving training data quality, incorporating retrieval-based augmentation, and applying post-hoc verification mechanisms \cite{lewis2020retrieval,manakul2023selfcheckgpt}. While these methods can reduce hallucination frequency to some extent, they often introduce additional computational overhead or rely heavily on external resources, making them difficult to generalize \cite{shuster2021retrieval}. Moreover, retrieval or verification alone does not explicitly address the internal reasoning structure of the model, which has been shown to be critical for reliable multi-step inference \cite{xiong2025deliberate}.

To address these limitations, this paper proposes a prompt-induced hallucination mitigation method based on an improved knowledge distillation chain-style model. By integrating structured knowledge and code-guided reasoning into the inference process, the proposed approach aims to enhance reasoning reliability while preserving the flexibility of large language models.

\section{Knowledge Distillation Chain-Style Model}

Knowledge distillation chain-style models combine knowledge distillation techniques with chain-of-thought reasoning to improve model interpretability and accuracy \cite{hinton2015distilling,wei2022chain}. This paradigm enables large language models to decompose complex tasks into intermediate reasoning steps, allowing the model to generate more coherent and logically consistent outputs \cite{kojima2022large}.

In a standard knowledge distillation chain-style framework, the model receives an input query and generates a sequence of reasoning steps before producing the final answer. These intermediate steps serve as an explicit reasoning trace, which helps guide the model toward the correct conclusion \cite{wang2023selfconsistency}. However, when the reasoning process itself relies solely on the model’s internal knowledge, errors may propagate across steps, leading to hallucinated conclusions \cite{ji2023survey}.

To alleviate this issue, we extend the knowledge distillation chain-style model by incorporating external structured knowledge such as knowledge graphs provide an explicit encoding of entities, relations, and temporal dependencies \cite{xiongtilp,xiong2024teilp}.
Specifically, the reasoning process is augmented with auxiliary information that constrains intermediate steps and reduces reliance on uncertain internal representations. 
This enhancement improves the model’s ability to maintain logical consistency across multiple reasoning stages \cite{nye2021show}.

\section{Improved Knowledge Distillation Chain with Code Guidance}

\subsection{Model Enhancement}

The improved knowledge distillation chain-style model introduces a code-guided module into the reasoning pipeline. This module is designed to guide knowledge exploration and reasoning by leveraging code as an explicit control mechanism. Instead of relying exclusively on natural language reasoning, the model uses code representations to constrain and direct the reasoning process.

The code module serves two primary purposes. First, it provides a structured mechanism for exploring relevant knowledge, enabling the model to systematically retrieve and reason over related concepts. Second, code is incorporated into the chain-of-thought prompts as an auxiliary representation, forming an external knowledge input that complements natural language reasoning.

By integrating code-guided reasoning, the model can better align intermediate steps with formal logic and structured knowledge, thereby reducing the likelihood of hallucinated reasoning paths.

\begin{figure}[t]
  \centering
  \includegraphics[width=0.5\linewidth]{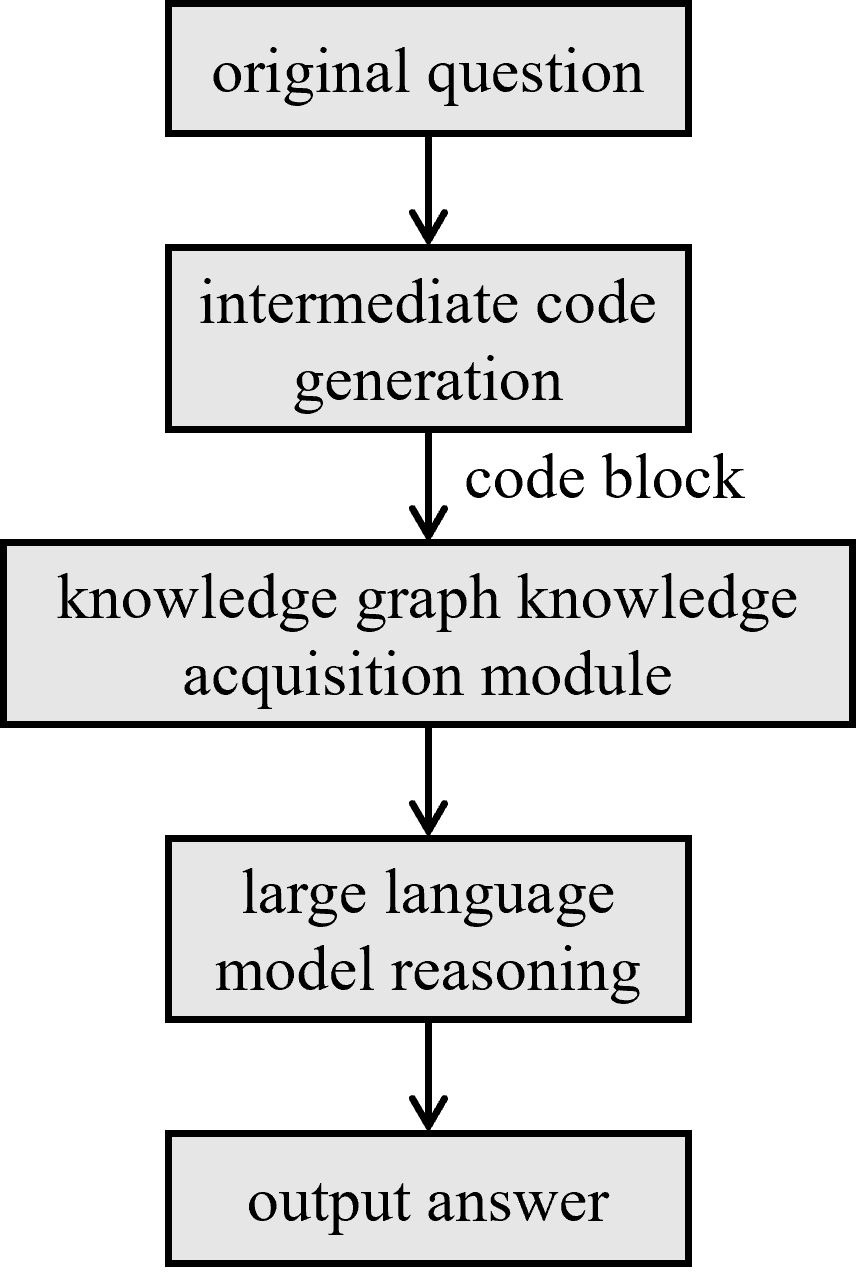}
  \caption{The process of suggesting hallucination problem-solving methods based on the large model based on the improved knowledge distillation chain.}
\end{figure}

\subsection{Reasoning Process Analysis}

Using the improved knowledge distillation chain-style model, we analyze the inference process of large language models. The explicit reasoning structure allows the model to verify intermediate conclusions and detect inconsistencies during inference. This process enhances the model’s ability to self-correct and improves overall answer accuracy.

Moreover, the explicit structure of the reasoning chain improves transparency and interpretability, making it easier to identify and diagnose sources of error in the model’s outputs.

\subsection{Prompt-Induced Hallucination Mitigation Method}

Based on the improved knowledge distillation chain-style model, we propose a prompt-induced hallucination mitigation method tailored for large language models. The method leverages structured reasoning and external knowledge guidance to reduce erroneous generation caused by ambiguous or incomplete prompts.

The mitigation process consists of three stages. First, the input prompt is analyzed and decomposed into structured sub-tasks. Second, the code-guided knowledge distillation chain generates intermediate reasoning steps under external constraints. Finally, the model produces a final answer that is grounded in both structured reasoning and validated intermediate steps.

This approach effectively reduces the propagation of reasoning errors and enhances the model’s robustness to prompt variations.

\begin{figure}[t]
  \centering
  \includegraphics[width=\linewidth]{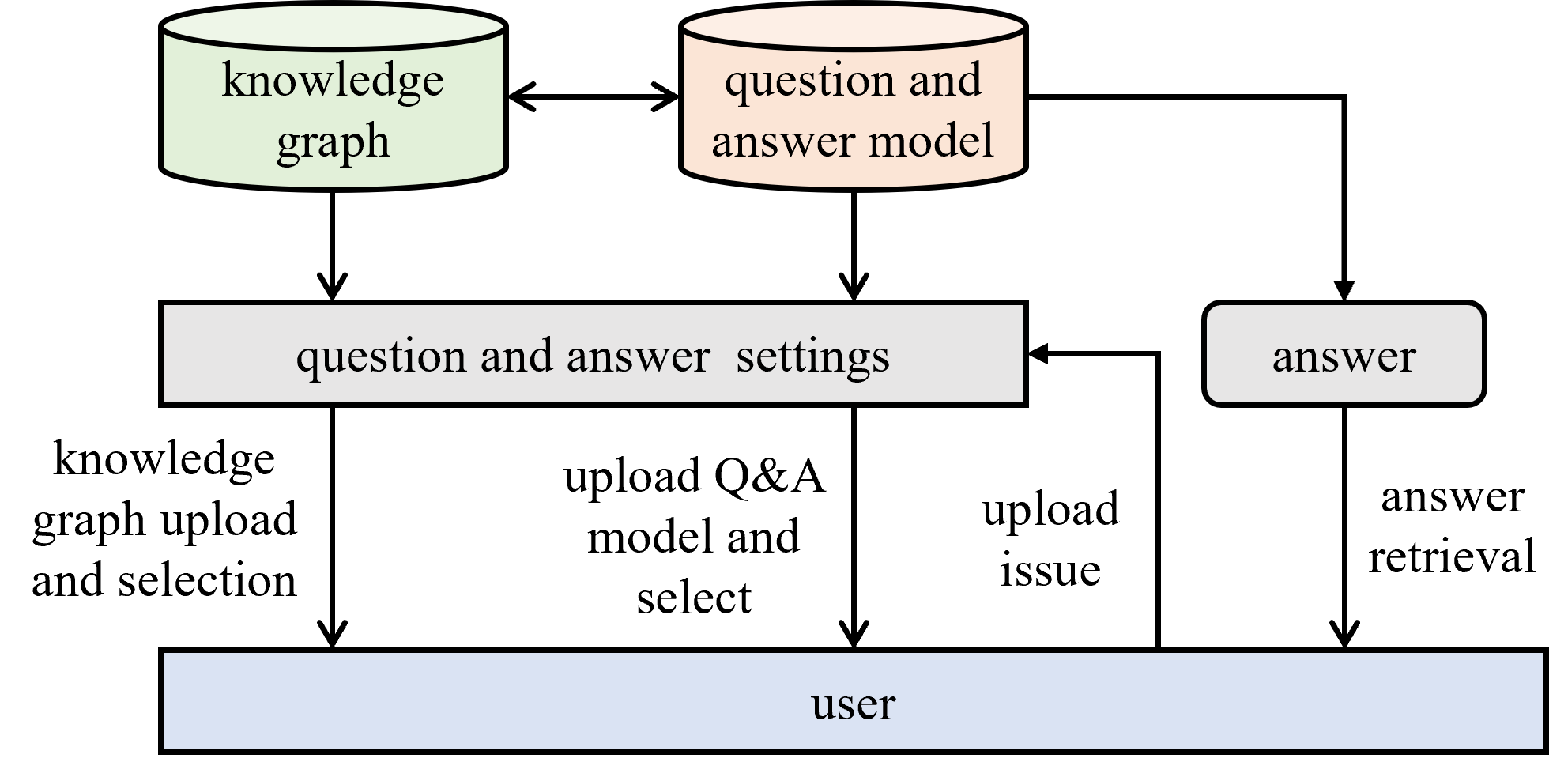}
  \caption{Simulation experiments.}
  \label{Fig5}
\end{figure}

\section{Experiments}

We conduct experiments on multiple public datasets to evaluate the effectiveness of the proposed method. Large language models such as GPT-4 and LLaMA~3.3 are used as base models for evaluation \cite{openai2023gpt4,touvron2023llama}. Performance is measured using standard information retrieval metrics, including HIT@1, HIT@3, and HIT@5 \cite{bordes2013translating}.

Experimental results demonstrate that introducing code-guided reasoning significantly improves the model’s contextual learning ability. Compared to baseline methods, the proposed approach achieves substantial performance gains across all evaluation metrics. In particular, the HIT@1, HIT@3, and HIT@5 scores exceed 95\%, indicating a strong reduction in prompt-induced hallucinations \cite{ji2023survey}.

These results confirm that the improved knowledge distillation chain-style model effectively enhances both accuracy and verifiability in large language model inference, consistent with prior findings on structured reasoning and external guidance \cite{nye2021show}.

\subsection{Experimental Setup}
To verify the effectiveness of the proposed method in addressing prompt-induced hallucination issues in large language models, a simulation experimental environment was constructed based on the Python programming language and the software tools OpenLink Virtuoso, OpenAI, and Treelib, and was executed on the Windows 10 operating system. The hardware configuration of the system is as follows: Intel(R) Core(TM) i7-8565U CPU, GeForce MX250, and 16 GB of memory. The simulation experimental results are shown in \ref{Fig5}.

\begin{table*}[t]
\centering
\caption{Improvement Verification Results of the Knowledge Distillation Chain Model (KDCM) (\%)}
\label{tab:improvement_verification}
\resizebox{0.7\linewidth}{!}{%
\begin{tabular}{l l c c c}
\toprule
\textbf{Dataset} & \textbf{Model} & \textbf{HIT@1} & \textbf{HIT@3} & \textbf{HIT@5} \\
\midrule
\multirow{2}{*}{WebQSP}
& KDCM
& 82.36 & 83.14 & 80.26 \\
& KDCM + Code Module
& 99.33 & 97.38 & 95.28 \\
\midrule
\multirow{2}{*}{CWQ}
& KDCM
& 81.36 & 82.09 & 82.14 \\
& KDCM + Code Module
& 97.86 & 98.03 & 96.20 \\
\midrule
\multirow{2}{*}{GSM8K}
& KDCM
& 82.06 & 85.79 & 84.39 \\
& KDCM + Code Module
& 98.23 & 95.14 & 95.47 \\
\midrule
\multirow{2}{*}{MWP}
& KDCM
& 85.26 & 84.39 & 82.11 \\
& KDCM + Code Module
& 98.19 & 96.78 & 95.08 \\
\midrule
\multirow{2}{*}{Dr.\ SPIDER}
& KDCM
& 86.29 & 83.14 & 85.29 \\
& KDCM + Code Module (Ours)
& 94.10 & 93.22 & 92.18 \\
\bottomrule
\end{tabular}
}
\end{table*}

\begin{figure*}[t]
  \centering
  \includegraphics[width=0.9\linewidth]{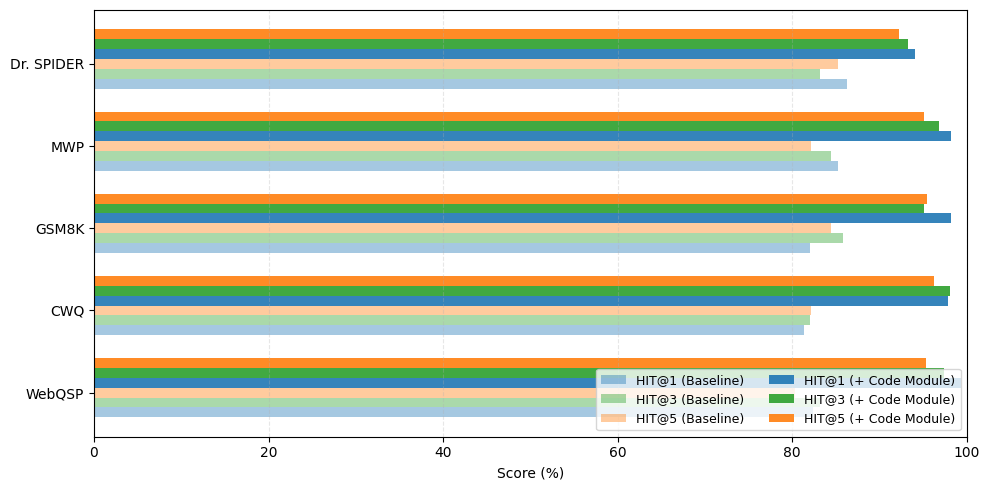}
  \caption{Verification results of the improvement of the knowledge distillation chain model.}
\end{figure*}

\subsection{Datasets and Preprocessing}
The experiments in this paper use publicly available datasets, including web-based question–answering datasets (WebQuestionsSP, WebQSP), CWQ (Complex Web Questions), GSM8K, MWP (Math Word Problems), and the Dr. SPIDER dataset, to evaluate the performance of the proposed method \citep{
yih2016value,
talmor2018web,
cobbe2021training,
koncel2015parsing,
li2023drspider,
berant2013semantic}.

The WebQSP dataset consists of a collection of question–answer pairs extracted from the Internet, covering multiple domains such as science and education, and contains a total of 4,737 question–answer pairs.

\begin{table}[t]
\centering
\caption{Robustness Verification Results (\%)}
\label{tab:robustness_verification}
\resizebox{0.85\linewidth}{!}{%
\begin{tabular}{l c c c}
\toprule
\textbf{Dataset} & \textbf{HIT@1} & \textbf{HIT@3} & \textbf{HIT@5} \\
\midrule
WebQSP     & 99.33 & 97.38 & 95.28 \\
CWQ        & 97.86 & 98.03 & 96.20 \\
GSM8K      & 98.23 & 95.14 & 95.47 \\
MWP        & 98.19 & 96.78 & 95.08 \\
Dr.\ SPIDER & 98.12 & 96.36 & 95.42 \\
\midrule
Average    & 98.35 & 96.74 & 95.49 \\
\bottomrule
\end{tabular}
}
\end{table}

The CWQ dataset includes two components: Question Files and Web Snippet Files, which contain 34,689 and 12,725,989 data instances, respectively.

The GSM8K dataset is a benchmark for evaluating AI systems in basic mathematics. It contains 8,500 high-quality multilingual elementary-level math problems, with 7,500 examples in the training set and 1,000 examples in the test set. Each data instance includes two fields.

The MWP dataset mainly consists of multi-step arithmetic and systems of linear equations, comprising one million data instances, and is commonly used to solve complex mathematical problems.

The Dr. SPIDER dataset is a large-scale dataset containing complex natural language data. It includes three sub-datasets focusing on data perturbation, structured query perturbation, and natural language question perturbation, and is primarily used to evaluate large language models’ capabilities in interpretability tasks and dialog reasoning.

\subsection{Evaluation Metrics}

To evaluate the effectiveness of the proposed method in mitigating prompt-induced hallucinations, we adopt HIT@K as the primary evaluation metric, which is widely used in information retrieval and knowledge-intensive reasoning tasks \cite{bordes2013translating}. HIT@K measures whether the correct answer appears within the top $K$ candidate responses generated by the model for a given query. A higher HIT@K score indicates stronger reasoning reliability and reduced hallucination behavior \cite{manakul2023selfcheckgpt}.

\begin{table}[t]
\centering
\caption{Mean Evaluation Metrics of Different Methods on Experimental Datasets (\%)}
\label{tab:mean_evaluation_methods}
\resizebox{0.95\linewidth}{!}{%
\begin{tabular}{l c c c}
\toprule
\textbf{Method} & \textbf{HIT@1} & \textbf{HIT@3} & \textbf{HIT@5} \\
\midrule
Average (Ours) & 98.40 & 96.83 & 95.51 \\
KG-LLM-PR & 91.06 & 91.78 & 90.22 \\
LLM-SubKG-Sum & 92.23 & 91.89 & 90.17 \\
RAG & 90.23 & 90.28 & 90.18 \\
Self-Check & 91.25 & 92.35 & 91.27 \\
\bottomrule
\end{tabular}
}
\end{table}

\begin{table}[t]
\centering
\caption{Generalization Verification Results (\%)}
\label{tab:generalization_verification}
\resizebox{0.95\linewidth}{!}{%
\begin{tabular}{l c c c}
\toprule
\textbf{Method} & \textbf{HIT@1} & \textbf{HIT@3} & \textbf{HIT@5} \\
\midrule
Proposed Method & 99.18 & 97.64 & 95.12 \\
KG-LLM-PR & 90.26 & 88.52 & 86.47 \\
LLM-SubKG-Sum & 92.36 & 90.11 & 86.25 \\
RAG & 90.36 & 90.25 & 91.09 \\
Self-Check & 90.28 & 91.41 & 91.26 \\
\bottomrule
\end{tabular}
}
\end{table}

\begin{figure}[t]
  \centering
  \includegraphics[width=\linewidth]{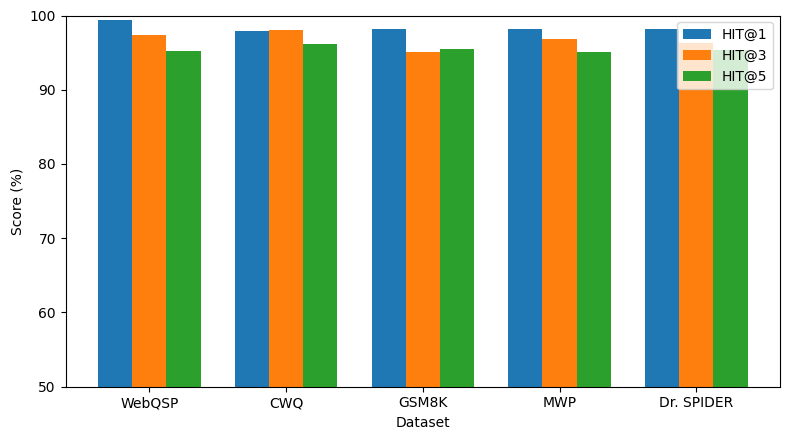}
  \caption{Robustness Verification Results.}
\end{figure}

\begin{figure}[t]
  \centering
  \includegraphics[width=\linewidth]{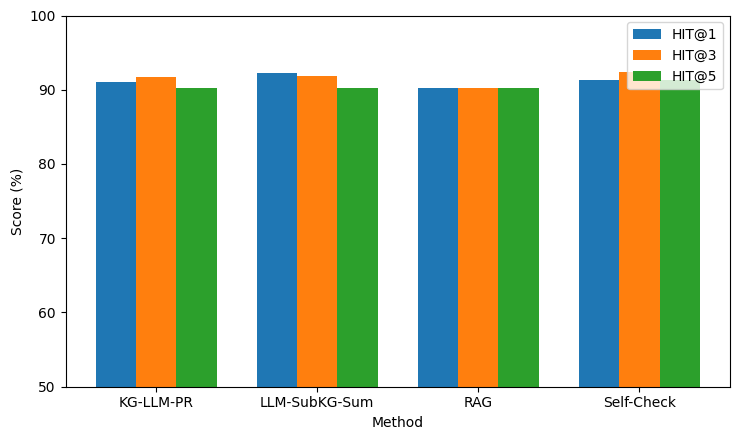}
  \caption{Mean Evaluation Indexes of Different Methods.}
\end{figure}

\begin{figure}[t]
  \centering
  \includegraphics[width=\linewidth]{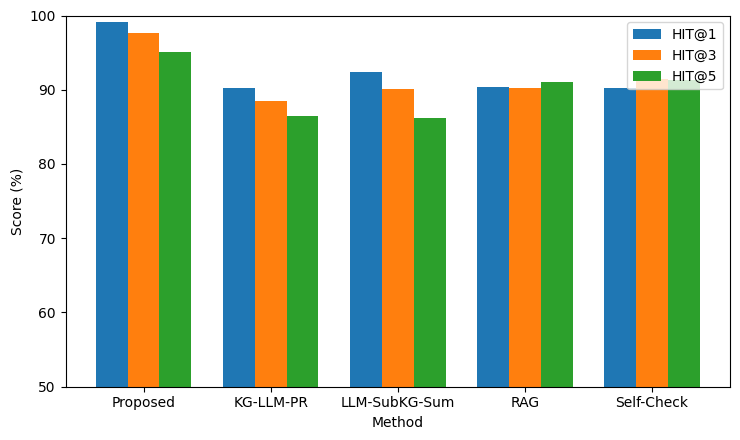}
  \caption{Generalization Verification Results.}
  \end{figure}

\begin{figure}[t]
  \centering
  \includegraphics[width=\linewidth]{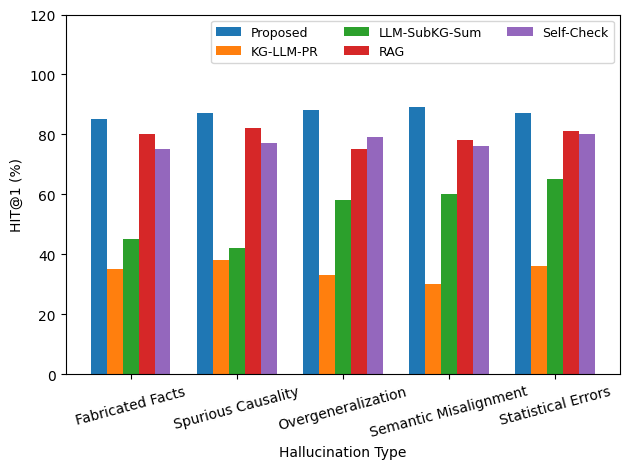}
  \caption{Results of the Proposed Method on Common Hallucination Types.}
\end{figure}

Formally, HIT@K is defined as:
\begin{equation}
\mathrm{HIT@K} = \frac{M}{N},
\end{equation}
where $N$ denotes the total number of test questions, and $M$ denotes the number of questions for which the correct answer is ranked within the top $K$ generated results.

\subsection{Results and Analysis}
We evaluate the proposed method using GPT-4 and LLaMA~3.3 as representative large language models \cite{openai2023gpt4,touvron2023llama}.
For each dataset, we compare the baseline large language model with its enhanced variant using the improved knowledge distillation chain-style model. 
We also compare with KG-LLM-PR \citep{zhang2025kgllm} and LLM-SubKG-Sum \citep{zhang2024kgentity}.
All models are evaluated under identical inference settings to ensure fairness. Performance is reported using HIT@1, HIT@3, and HIT@5 metrics.

Experimental results demonstrate that the proposed method consistently improves performance across all evaluated datasets. Compared to baseline models, the improved knowledge distillation chain-style model achieves substantial gains in HIT@1, HIT@3, and HIT@5, indicating a significant reduction in prompt-induced hallucinations.

The results show that incorporating code-guided reasoning enhances the model’s ability to learn and utilize contextual information effectively. By constraining intermediate reasoning steps with structured knowledge, the model reduces reliance on uncertain internal representations and produces more accurate and verifiable outputs.

Across different datasets, the proposed method exhibits stable improvements, suggesting strong generalization capability. Notably, performance gains are particularly pronounced in tasks that require multi-step reasoning, where hallucination errors are more likely to accumulate in standard chain-of-thought reasoning.

\subsection{Robustness Analysis}

To assess robustness, we evaluate the proposed method under variations in prompt formulation and dataset characteristics. The results indicate that the improved model maintains high HIT@K performance even when input prompts are ambiguous or incomplete.

This robustness can be attributed to the structured reasoning process enforced by the improved knowledge distillation chain-style model. By explicitly guiding intermediate reasoning steps, the model becomes less sensitive to prompt noise and reduces error propagation during inference.

\subsection{Generalization Evaluation}

We further examine the generalization ability of the proposed method by evaluating it on datasets that differ from those used during model tuning. The results demonstrate that the method generalizes well across domains, maintaining high accuracy and low hallucination rates.

Compared with existing hallucination mitigation approaches, the proposed method achieves superior performance across multiple evaluation settings. This indicates that the method does not rely on dataset-specific heuristics and can be effectively applied to a wide range of large language model applications.

\section{Conclusion}

This paper presents a prompt-induced hallucination mitigation method based on an improved knowledge distillation chain-style model for large language models. By incorporating code-guided reasoning and structured external knowledge into the inference process, the proposed approach improves reasoning accuracy, robustness, and interpretability. Experiments on multiple public datasets demonstrate consistent performance gains, with HIT@1, HIT@3, and HIT@5 exceeding 95\% in several settings, indicating a substantial reduction in hallucination behavior while maintaining model flexibility. Future work will investigate extending the framework to multimodal reasoning tasks and integrating it with retrieval-augmented generation and reinforcement learning-based optimization techniques.

\section*{Limitations}

Despite its effectiveness, the proposed method has several limitations. First, the introduction of code-guided reasoning increases inference complexity and may lead to higher computational overhead compared to standard prompt-based approaches. Second, the method relies on the availability of well-structured external knowledge and task-appropriate code representations, which may limit its applicability in domains where such resources are scarce. Finally, while the approach demonstrates strong performance on text-based reasoning benchmarks, its effectiveness in fully open-ended generation and multimodal scenarios remains to be validated.

% Bibliography entries for the entire Anthology, followed by custom entries
%\bibliography{anthology,custom}
% Custom bibliography entries only
\bibliography{custom}

\end{document}